\documentclass[letterpaper, 10 pt, journal, twoside]{IEEEtran}

\IEEEoverridecommandlockouts                              

\usepackage{amsmath} 
\usepackage{amssymb}
\usepackage{multirow}
\usepackage{graphicx}
\usepackage{color}
\usepackage{booktabs}
\usepackage{soul}
\usepackage[switch]{lineno}
\usepackage{tabularx}
\usepackage[flushleft]{threeparttable}
\usepackage{booktabs}
\usepackage{url}
\usepackage{subfigure}
\usepackage{cite}
\usepackage{makecell}
\usepackage[switch]{lineno}
\usepackage{siunitx} 
\usepackage{stfloats}
\usepackage[table]{xcolor}
\usepackage{caption}
\soulregister\cite7
\soulregister\ref7
\title{DiffMap: Enhancing Map Segmentation with \\ Map Prior Using Diffusion Model
}

\author{Peijin Jia$^{1}$, Tuopu Wen$^{1}$, Ziang Luo$^{1}$, Mengmeng Yang$^{1}$, Kun Jiang$^{1}$, Ziyuan Liu$^{1}$,\\ Xuewei Tang$^{1}$,  Zhiquan Lei$^{1}$, Le Cui$^{2}$, Bo Zhang$^{2}$, Long Huang$^{3}$, Diange Yang$^{1}$
\thanks{This work was supported in part by the National Natural Science Foundation of China(U22A20104,52102464), Beijing Natural Science Foundation (L231008,L243008), Young Elite Scientist Sponsorship Program By BAST (BYESS2022153) and the State-funded postdoctoral researcher program of China (GZC20231285). This work was also sponsored by Tsinghua University-Daimler Joint Research Center for Future Mobility and Tsinghua University-Didi Joint Research Center.}
\thanks{$^{1}$Peijin Jia, Ziang Luo, Kun Jiang, Zhiquan Lei, Xuewei Tang, Ziyuan Liu, Tuopu Wen, Mengmeng Yang, Diange Yang are with the School of Vehicle and Mobility, Tsinghua University, Beijing, 100084, China. $^{2}$Le Cui and Bo Zhang are with DiDi Chuxing. $^{3}$Huang Long serves as the deputy director at Map Supervision Centre, Ministry of Natural Resources.}
\thanks{$^{*}$Primarily contact to jpj22@mails.tsinghua.edu.cn.}
\thanks{Corresponding author: Diange Yang, Kun Jiang and Mengmeng Yang}
\thanks{Digital Object Identifier (DOI): see top of this page.}}

\begin{document}
\maketitle

\begin{abstract}


Constructing high-definition (HD) maps is a crucial requirement for enabling autonomous driving. In recent years, several map segmentation algorithms have been developed to address this need, leveraging advancements in Bird's-Eye View (BEV) perception. However, existing models still encounter challenges in producing realistic and consistent semantic map layouts. A prominent issue is the limited utilization of structured priors inherent in map segmentation masks.
In light of this, we propose DiffMap, a novel approach specifically designed to model the structured priors of map segmentation masks using latent diffusion model. By incorporating this technique, the performance of existing semantic segmentation methods can be significantly enhanced and certain structural errors present in the segmentation outputs can be effectively rectified. Notably, the proposed module can be seamlessly integrated into any map segmentation model, thereby augmenting its capability to accurately delineate semantic information.
Furthermore, through extensive visualization analysis, our model demonstrates superior proficiency in generating results that more accurately reflect real-world map layouts, further validating its efficacy in improving the quality of the generated maps. 

\end{abstract}

\section{Introduction}
\begin{figure}[t]
    \centering
    \includegraphics[width=0.45\textwidth]{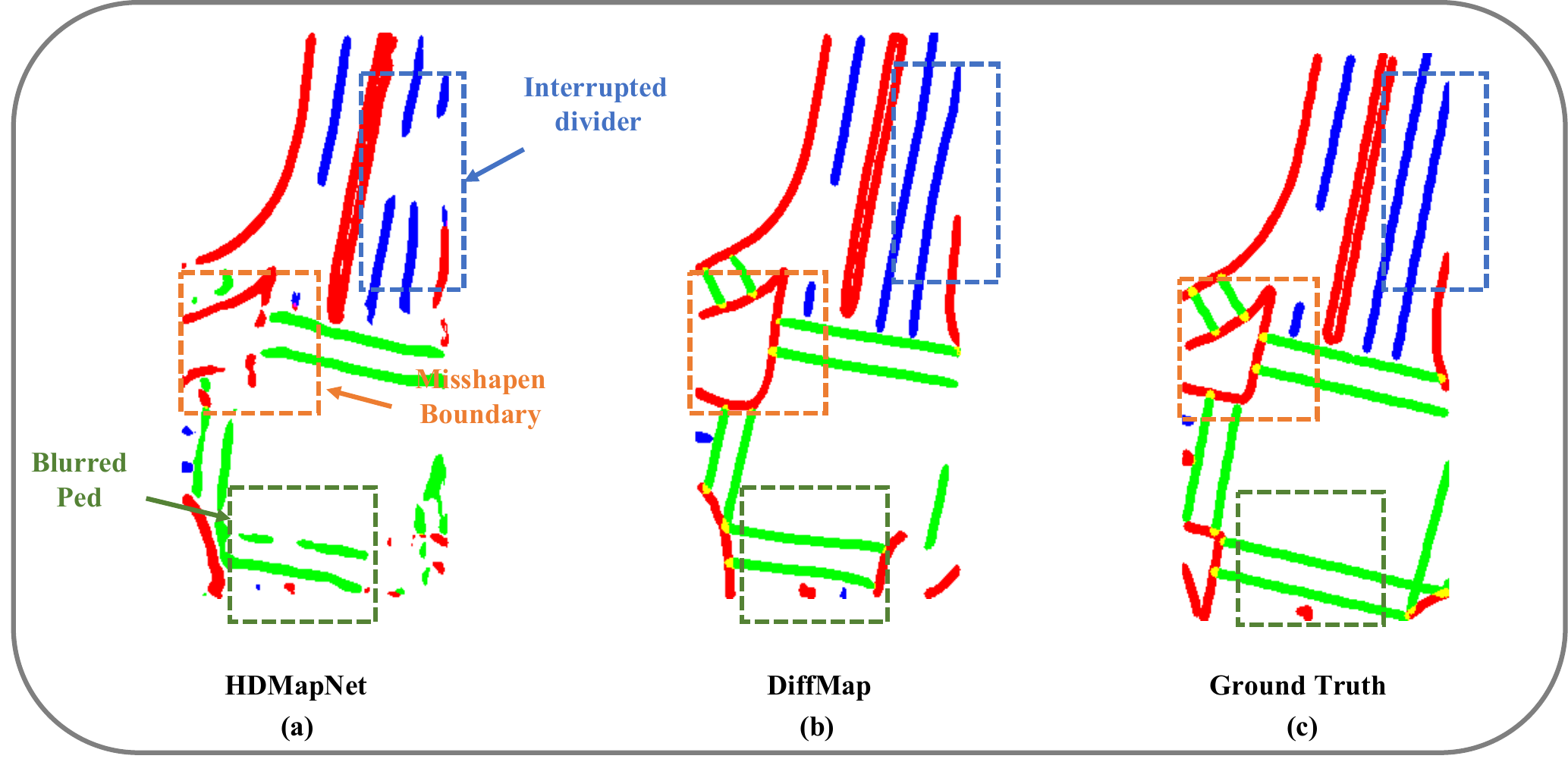}
    \caption{(a) shows the problems of the traditional bev map segmentation model, (b) shows the prediction result of our model which effectively corrects the previous structural problems and is closer to the ground truth in (c).}
    \label{fig:bug}
\end{figure}

\IEEEPARstart{H}{D} maps are essential for facilitating accurate environmental understanding and precise navigation in autonomous vehicles. However, the traditional generation process of these maps is a laborious and complex process. To address this challenge, the integration of map construction into the BEV (Bird's-Eye View) perception task has gained considerable attention.
Current research regards the construction of rasterized HD maps as a segmentation task in the BEV space, primarily involving architectures similar to Fully Convolutional Networks (FCNs)\cite{long2015fully} with segmentation heads after obtaining the BEV features. For instance, in short-range perception, HDMapNet\cite{li2022hdmapnet} encodes sensor features through LSS\cite{philion2020lift} and then employs multi-branch FCNs for semantic segmentation, instance detection, and direction prediction to construct the map. For long-range map construction, SuperFusion\cite{dong2022superfusion} follows the similar pipeline with camera and LiDAR fusion mechanisms. Despite ongoing improvements in map segmentation, the pixel-based classification approach of segmentation exhibits inherent limitations, including the potential neglect of specific category attributes, which can lead to issues such as distorted and interrupted dividers, blurred pedestrian crossings, and other types of artifacts and noise, as shown in Fig.\ref{fig:bug} (a). These problems not only affect the structural accuracy of the map but could also directly impact downstream path planning modules of autonomous driving systems.

Hence, it is desirable for the model to consider the structural prior information of HD maps such as the parallel and straight nature of lane lines. However, capturing such properties of HD map elements proves challenging when optimizing the classification loss in discriminative segmentation models. 
Conversely, generative models have exhibited their ability to capture the realism and inherent characteristics of images.
Notably, the LDM (Latent Diffusion Model)\cite{rombach2022high} has demonstrated significant potential in high fidelity image generation and has proven effective in tasks related to segmentation enhancement\cite{lai2023denoising},\cite{baranchuk2021label}. By incorporating control variables\cite{zhang2023adding}, the generation of images can be further guided to meet specific control requirements. As a result, applying the generative model to capture the map structural prior has great potential to reduce the segmentation artifact and improve in the map construction performance. 

Therefore, we propose a network named DiffMap for online map reconstruction task.
The network uses a modified LDM as an enhancement module to model map structured prior for existing segmentation models. 
To migrate the diffusion technique into the online mapping task, there are two main challenges need to be resolved. 
First, the mapping results are highly determined by online sensor observations. 
As a result, we design a condition injection module that integrates the online BEV feature of sensor observations into the denoising process as control signals. 
Second, since downstream task usually requires instance-wise mapping result, to enable the diffusion network to reconstruct the map elements separately, an additional discriminative module is proposed to help the network learn the distinguishable features of different map elements.
Our experimental results demonstrate that DiffMap can effectively generate a more smooth and reasonable map segmentation while greatly reducing the artifacts, therefore improving the overall map construction performance.
This is the first study to incorporate a LDM model into map prior modeling and be used as a plug-and-play module to enhance any map segmentation task. 
We plan to share all our code and data to facilitate further research and application in the community.

\section{Related Works}
\subsection{Semantic map construction}
In traditional HD maps construction, semantic maps are typically manually or semi-automatically annotated based on the LiDAR point clouds. This involves using SLAM-based algorithms\cite{shan2018lego, shan2020lio} to build globally consistent maps, with semantic annotations manually added to the maps. However, this approach is time-consuming and labor-intensive, and it presents challenges in terms of updating the maps, thereby limiting scalability and real-time performance. HDMapNet\cite{li2022hdmapnet} pioneered a method for dynamically constructing local semantic maps using onboard sensors. It encodes LiDAR point clouds and surround-view image features into BEV space, and decodes with three different heads. Finally produce a vectorized local semantic map with post-processing.
SuperFusion\cite{dong2022superfusion}, in contrast, focuses on constructing long-range high-precision semantic maps, leveraging LiDAR depth for enhancing image depth estimation and employing image features to guide remote LiDAR feature prediction. And then employs a map detection head akin to HDMapNet to get semantic maps. 
MachMap\cite{qiao2023machmap} divides the task into
line detection of polylines and instance segmentation of polygons and uses post-processing to refine the mask to obtain the final result.
The subsequent work\cite{liu2023vectormapnet, liao2022maptr, liao2023maptrv2, xu2023insightmapper} explores end-to-end online mapping and gets the direct vectorized HD map.
Dynamic construction of semantic maps without manual annotation effectively reduces the construction cost. 

However, due to potential issues such as occlusion and malfunction of onboard sensors, coupled with susceptibility to adverse weather conditions, their uncertainty poses a series of safety concerns for autonomous vehicles. This significantly hampers the widespread adoption and application of semantic map segmentation tasks.

\subsection{Diffusion model for segmentation and detection}
Denoising Diffusion Probabilistic Models (DDPMs) \cite{ho2020denoising} are a class of generative models based on Markov chains, which have demonstrated outstanding performance in fields such as image generation. Nowadays, DDPMs have gradually expanded their application to various tasks including segmentation and detection. SegDiff\cite{amit2021segdiff} applies the diffusion model to address image segmentation tasks, where the UNet encoder used in the diffusion model is further decoupled into three modules: E, F, and G. Modules G and F encode the input image $\mathbf{I}$ and segmentation map $\mathbf{X}_t$, respectively, which are then merged by addition in E to iteratively refine the segmentation map. DDPMS\cite{lai2023denoising} produce an initial prediction prior with base segmentation models, and refines the  
 prior with diffusion model. DiffusionDet\cite{chen2023diffusiondet} extends the diffusion model to the object detection framework, modeling object detection as a denoising diffusion process from noisy boxes to object boxes. Diffusion is also utilized in the field of autonomous driving, such as in the case of MagicDrive\cite{gao2023magicdrive}, where it is applied in the synthesis of street views guided by geometric constraints and Motiondiffuser\cite{jiang2023motiondiffuser} extends the diffusion model to the multi-agent motion prediction problems. 

 The above work demonstrates the scalability of the diffusion model in segmentation, detection, and other fields. In semantic maps construction, the generative effect of the diffusion model can effectively complement the uncertainty introduced by sensor information. Yet, there has been no prior work applying the diffusion model to semantic map construction. 

\subsection{Map prior}
There are currently several methods to enhance model robustness and reduce uncertainty in onboard sensors by leveraging prior information, including explicit standard map information and implicit temporal information. MapLite2.0\cite{ort2022maplite} utilizes a Standard Definition (SD) prior map as a starting point and incorporates onboard sensors to infer the local HD map in real time. MapEx\cite{sun2023mind} and SMERF\cite{luo2023augmenting} utilize standard map data to improve lane perception and topological understanding. SMERF\cite{luo2023augmenting} employs a Transformer-based standard map encoder to encode lane lines and lane types, and then computes cross-attention between the standard map information and sensor-based bird's-eye view (BEV) features, integrating the standard map information. NMP\cite{xiong2023neural} provides long-term memory capabilities for autonomous vehicles by integrating past map prior data with current perception data. MapPrior\cite{zhu2023mapprior} combines discriminative and generative models, where preliminary predictions are generated based on existing models and encoded as priors into the discrete latent space of the generative model during the prediction stage. The generative model is then used to refine the predictions. PreSight\cite{yuan2024presight} involves optimizing a city-scale neural radiance field using data from previous journeys to generate neural priors, enhancing online perception in subsequent navigations.

While the mentioned methods can improve the performance of map segmentation networks to some extent, there is a lack of research specifically focused on enhancing the prior structure of maps. Further exploration in this area is necessary.

\section{Methodology}
\label{method}
\subsection{Preliminaries}

DDPMs are a type of generative models that leverage diffusion processes to approximate the underlying data distribution $p(\mathbf{x})$. They function by systematically removing noise from a variable that is normally distributed, echoing the reverse operation of a Markov Chain of predetermined length $T$. Within DDPMs, the transformation from a clean initial data point $\mathbf{x}_0$ to a noisy one $\mathbf{x}_T$ is accomplished through a series of conditional probabilities $ q(\mathbf{x}_t | \mathbf{x}_{t-1})$, with $q$ denoting the noise introduction process. At each step $t$, noise is added according to the following specification:

\begin{equation}
\label{ddpm}
    q(\mathbf{x}_t | \mathbf{x}_{t-1}) = \mathcal{N}(\mathbf{x}_t; \sqrt{1 - \beta_t} \mathbf{x}_{t-1}, \beta_t \mathbf{I})
\end{equation}
The term \( \beta_t \) is the regulator of the noise level, \( \mathcal{N} \) signifies a normal distribution, and \( \mathbf{I} \) is the identity matrix. The final result of this progression is that \( \mathbf{x}_T \) becomes indistinct from random noise. The subsequent denoising mechanism is an iterative reconstruction from \( \mathbf{x}_T \) to \( \mathbf{x}_0 \), driven by a denoising network \( \mathbf{\epsilon}_\theta \) that estimates and progressively refines the clean data points from the noise. The denoising function is described by the equation:

\begin{equation}
   p_\theta(\mathbf{x}_{t-1} | \mathbf{x}_t) = \mathcal{N}(\mathbf{x}_{t-1}; \mu_\theta(\mathbf{x}_t, t), \Sigma_\theta(\mathbf{x}_t, t))
\end{equation}
Here, $\mu_\theta$ and $\Sigma_\theta$ denote the mean and covariance estimated by the denoising network $\epsilon_\theta$, respectively.

In a similar vein, Latent Diffusion Models (LDMs) adopt an incremental denoising approach yet operate within a more compact and efficient latent space. This reduced space strips away high-frequency details that are typically imperceptible, allowing the model to focus on capturing the core semantic features of the data. 
This efficiency makes the latent space a more fitting arena for training likelihood-based generative models, as opposed to the computationally demanding high-dimensional pixel space. The objective function for LDMs is articulated as follows:

\begin{equation}
    L_{\text{LDM}} = \mathbb{E}_{\mathbf{x}, \mathbf{\epsilon} \sim \mathcal{N}(0, 1), t} \left[ \| \epsilon - \epsilon_\theta(\mathbf{z}_t, t) \|^2 \right]
\end{equation}
where $\mathbf{z}_t=D(\mathbf{x}_t)$ is the corresponding data point of $\mathbf{x}_t$ in the latent space. The transformation $D$ is modeled as a feedforward neural network.
The denoising process of LDMs $\epsilon_\theta$ is instantiated as a time-conditioned UNet applied on $\mathbf{z}_t$.
This approach not only enhances the image synthesis efficiency but also ensures the production of high-fidelity images with significantly lower computational costs during both training and inference stages.

Besides, the integration of a conditional denoising function \( \epsilon_{\theta}(\textbf{z}_t, t, \textbf{y}) \) within these models allows for the controlled synthesis process when influenced by various input forms \( \textbf{y} \), including textual annotations, semantic maps, and the broader domain of image-to-image translations. The conditional LDM is learned through a process that focuses on the interplay between the conditioning input and the latent representation, which is reflected in the objective function:

\begin{equation}
    L_{\text{LDM}} = \mathbb{E}_{\mathbf{x},\mathbf{y},\epsilon \sim \mathcal{N}(0,1),t} \left[ \left\| \epsilon - \epsilon_{\theta}(\mathbf{z}_t, t, \tau_{\theta}(\mathbf{y})) \right\|^2 \right]
\end{equation}
where \( \tau_{\theta} \) is a domain specific encoder that projects condition $\mathbf{y}$ to an intermediate representation.
This configuration ensures that \( \tau_{\theta} \) and \( \epsilon_{\theta} \) are jointly optimized, showcasing the adaptability of the conditioning mechanism. Such a system can be enriched with specialized knowledge for various domains, for example, by incorporating transformer models for text-based prompts. 

\subsection{Architecture}
\begin{figure*}[!ht]
    \centering
    \includegraphics[width=0.87\textwidth]{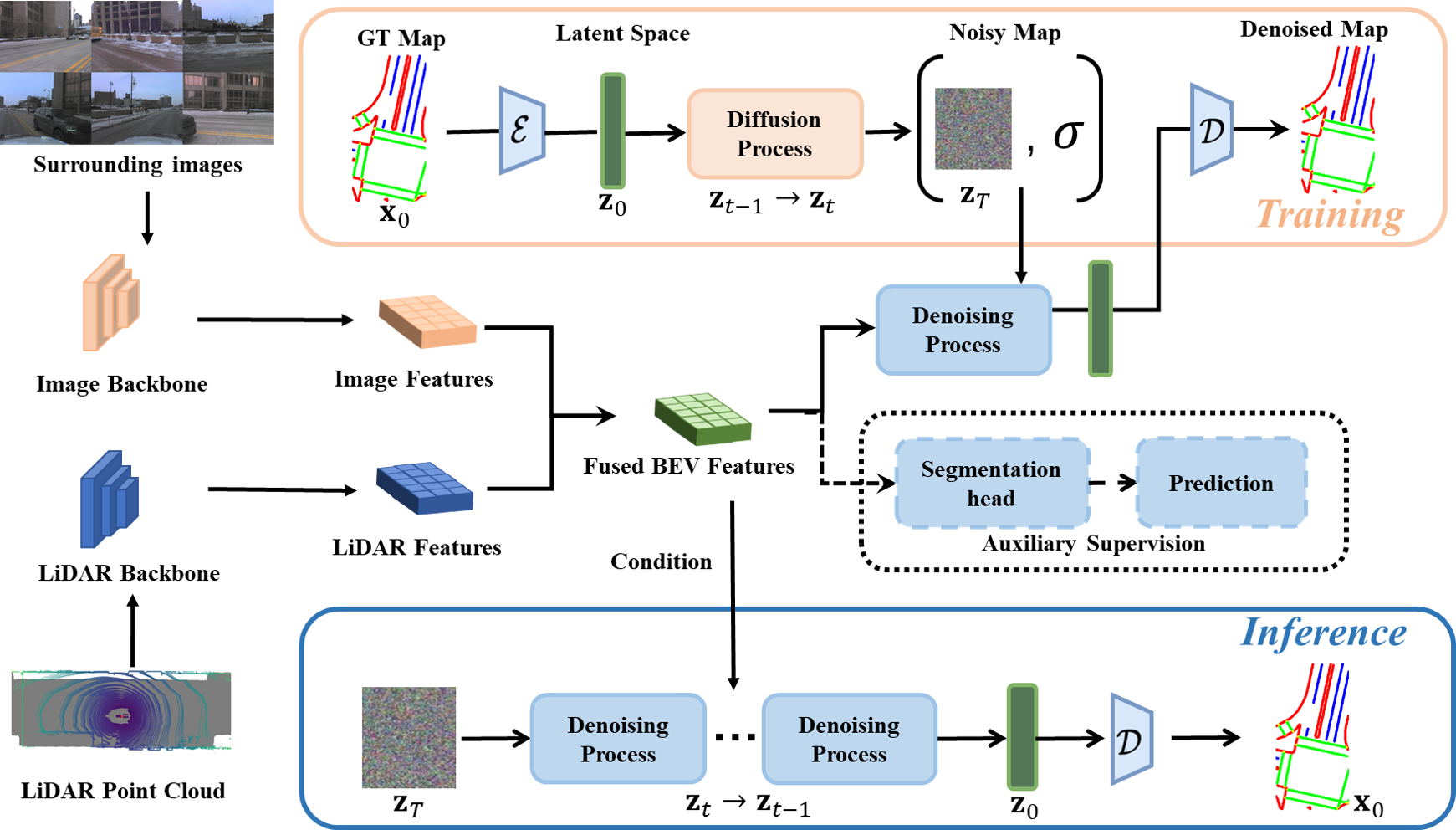}
    \caption{\textbf{Architecture Overview:} After extracting features from surrounding multi-view images and LiDAR point clouds separately using backbone networks, the features are transformed into Bird's Eye View (BEV) space for fusion. During the training process, random noise is continuously added to the ground truth map. Then, in the denoising process, the fused BEV features are used as conditional control variables of Diffmap, ultimately generating the predicted segmentation map. Whereas in the inference process, results are obtained in the continuous denoising from random noise.}
    \label{fig:architecture}
\end{figure*}

We propose DiffMap as a decoder to incorporate the diffusion model into semantic map segmentation models. 
The overall framework of DiffMap is illustrated in Fig. \ref{fig:architecture}. 
The model takes surrounding multi-view images and LiDAR point clouds as inputs, encodes them into BEV space and gets the fused BEV features. 
DiffMap is then employed as the decoder to generate segmentation maps. 
Within the DiffMap module, we utilize the BEV features as conditions to guide the denoising process. In this section, we first introduce the base model for generating BEV features, followed by an explanation of DiffMap and its denoising module. 
Finally, we present the training and inference processes.

\subsubsection{Baseline for Semantic map construction}

The baseline adopted in our work primarily follows a BEV encoder-decoder paradigm. The encoder part is responsible for extracting features from the input data (LiDAR and/or camera data), transforming them into high-dimensional representations. Meanwhile, the decoder, typically serving as the segmentation head, maps the high-dimensional feature representations to corresponding segmentation maps.

The baseline plays two main roles throughout the framework: supervisor and controller. 
As a supervisor, the baseline generates segmentation results as auxiliary supervision. 
Simultaneously, as a controller, it provides the intermediate BEV features  $\mathcal{B} \in \mathbb{R}^{H \times W \times C}$ as conditional control variables to guide the generation process of the diffusion model.

\subsubsection{DiffMap Module}

Following LDM, we introduce the DiffMap module as the decoder within our baseline framework. LDM mainly consists of two parts: an image perceptual compression module such as VQVAE\cite{oord2018neural} and a diffusion model constructed using UNet\cite{ronneberger2015unet}. Initially, the encoder $\varepsilon$ encodes the map segmentation ground truth $\mathbf{x}$ into the latent space as $\mathbf{z}={\varepsilon}(\mathbf{x}) \in \mathbb{R}^{H' \times W' \times D'}$, where $D'$ represents the low dimensions of the latent space. 
Subsequently, diffusion and denoising are performed in the low-dimensional latent variable space, and then the latent space is restored to the original pixel space using a decoder $\mathcal{D}$. 

Firstly, We add noise $\sigma$ through diffusion process and get the noisy latent map $\left\{\mathbf{z}_t\right\}_{t=0}^T$ at each timestep $t$, where $\mathbf{z}_0 = \mathbf{z}$.
Then during the denoising process, UNet serves as the backbone network for the noise prediction. 
Given that our primary objective is to utilize the diffusion model to refine map segmentation results, we hope to enhance the supervision of the segmentation results and expect the DiffMap model to directly provide semantic features for instance predictions during training. 

\begin{figure}[h]
    \centering
    \includegraphics[width=0.45\textwidth]{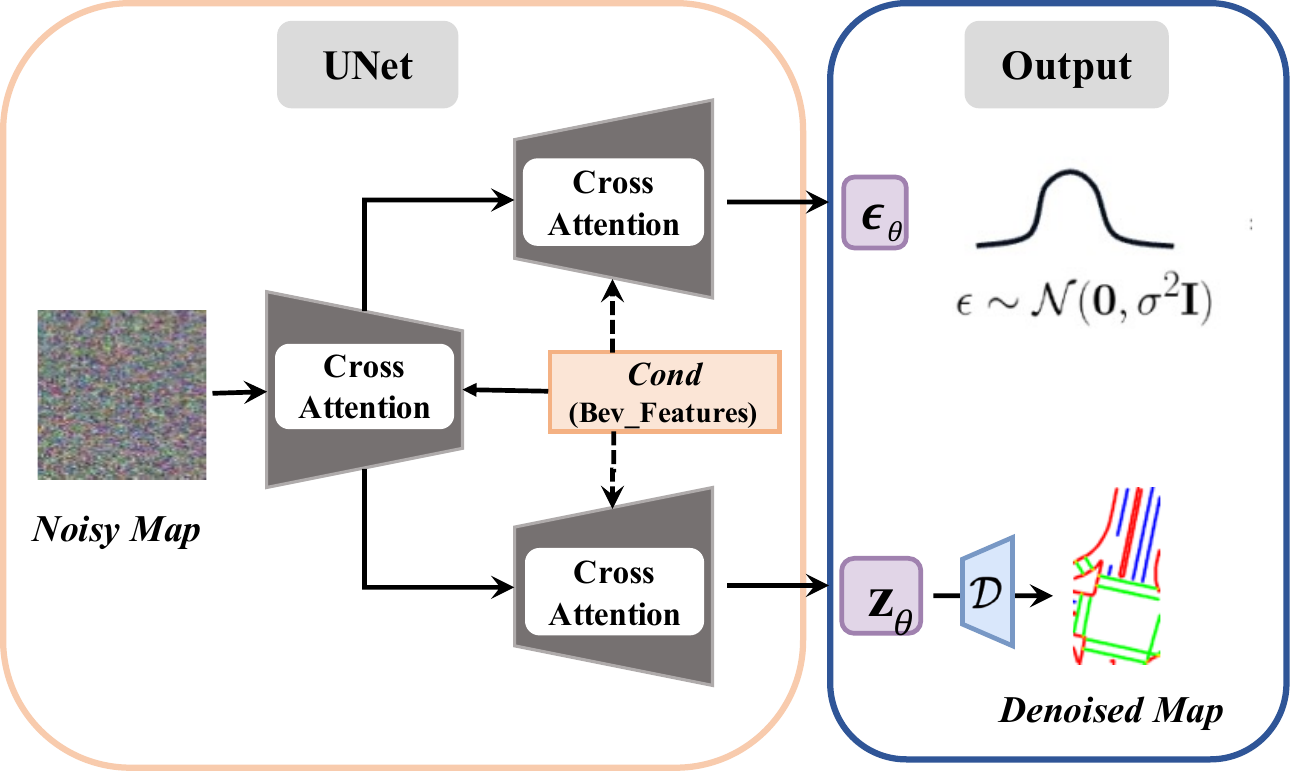}
    \caption{\textbf{Denoising Module:} In order to get a direct result of the segmented features, we decouple the UNet decoder into two branches, one branch is to predict noise $\epsilon$ as traditional diffusion model and the other is to predict the $\mathbf{z}$ in latent space. During the denoising process, we resize BEV features into latent space size as a conditional control variable. 
    We first concatenate it with noisy latent map, and then incorporate it into the two decoders of UNet with cross attention mechanism.}
    \label{fig:Denoising}
\end{figure}

Thus, we further decouple the UNet network structure into two branches like \cite{huang2023decoupled}, one branch is to predict noise $\epsilon$ as traditional diffusion model and the other is to predict $\mathbf{z}$ in the latent space.
The overall denoising module is illustrated in Fig. \ref{fig:Denoising}. 

As we want to obtain the map segmentation results under the current sensor input, this process should be a conditional generation process. The probability distribution we expect to obtain can be modeled as $p_{\theta}(\mathbf{x}|\mathbf{y})$, where $\mathbf{x}$ presents the map segmentation results and $\mathbf{y}$ denotes the conditional control variables, BEV features.
To achieve conditional generation, we incorporate the control variables in two ways.
Firstly, as $\mathbf{z}_{t}$ and BEV fatures $\mathcal{B}$ have the same category and scale in spatial domain, we can resize $\mathcal{B}$ into latent space size and then concatenate them as the input of denoising process. 
\begin{equation}
    \mathbf{z}_{t}' = \text{concat} (\mathbf{z}_{t}, \mathcal{B})
\end{equation}
Secondly, we incorporate a cross attention mechanism into each layer of UNet, where $\mathcal{B}$ plays as key/value and $\mathbf{z}_{t}$ plays as query, as shown in Fig. \ref{fig:Denoising}. The formulation of the cross attention module is illustrated as:
\begin{equation}
    \text{Attention}(\mathbf{z}_t', \mathcal{B}, \mathcal{B}) = \operatorname{softmax}\left(\frac{\mathbf{z}_t' \mathcal{B}^T}{\sqrt{d}}\right) \cdot \mathcal{B}
\end{equation}

\subsubsection{Instance Segmentation}
After getting the latent map prediction $\mathbf{z}_\theta$, we decode it to the original pixel space as the semantic feature map. 
Then we can obtain the instance prediction from it following the method proposed by HDMapNet, which outputs three predictions with different heads composed by MLPs: semantic segmentation, instance embedding, and lane direction. 
These predictions are subsequently used in the post-processing step to vectorize the map.

\subsection{Implementation}

\subsubsection{Training}
\label{train}
Firstly, we train VQVAE to compress the original segmentation ground truth into latent space. 
During training, the total training objective becomes:
\begin{equation}
\begin{aligned}
L=\underbrace{\log p\left(\mathbf{x} | z_q(\mathbf{x})\right)}_{\text {reconstruction loss }}&+\underbrace{\left\|\operatorname{sg}\left[z_e(\mathbf{x})\right]-e_k\right\|_2^2}_{\text {VQ loss }}\\
&+\underbrace{\beta\left\|z_e(\mathbf{x})-\operatorname{sg}[e]\right\|_2^2}_{\text {commitment loss}} 
\end{aligned}
\end{equation}
where $z_e(\mathbf{x})$ stands for the output of the encoder, $e$ stands for the embedding space, $z_q(\mathbf{x})$ stands for the the nearest embedding of $z_e(\mathbf{x})$ in the embedding space, and $sg(\cdot)$ is the stop-gradient operator. 
Through VQVAE, the ground truth maps can be compressed into the latent space, improving generation efficiency and preventing the diffusion model from over-training on image pixels.

During the training of the diffusion model, our objective is to minimize the reconstruction distance between the predicted images (or noise) and their ground truth counterparts. Since we decouple the UNet branches, the training objective is formulated as follows:
\begin{equation}
\mathop{\min}\limits_{\theta}
\mathbb{E}_{q(\mathbf{x}_0)} \mathbb{E}_{q(\epsilon)} \left( \left\| {\mathbf{z}}_{\theta} - \mathbf{z} \right\|^2 + \left\|{\epsilon}_{\theta} - {\epsilon} \right\|^2 \right)
\end{equation}
In the equation, ${\mathbf{z}}_{\theta}$ and ${\epsilon}_{\theta}$ correspond to the image branch and noise branch, respectively. Subsequently, the model is trained to predict images and noise separately. 
Additionally, to achieve higher-level image generation accuracy, we also incorporate the losses from baseline model as auxiliary supervision, including the cross-entropy loss for the semantic segmentation, and the discriminative loss for the instance embedding mentioned in HDMapNet. 
The final objective function combines the loss of diffusion model $\mathcal{L}_\text{diff}$ with the auxiliary losses from the baseline model $\mathcal{L}_\text{baseline}$, resulting in the following mathematical expression:
\begin{equation}
\mathcal{L} = {\mathcal L}_{\text{diff}} + {\mathcal L}_{\text{baseline}} 
\end{equation}

\subsubsection{Inference}
During the inference stage, we start with a noisy image and iteratively denoise it based on the parameters obtained from training. After multiple denoising iterations, we get $\mathbf{z}_0$, in which the noise becomes negligible. Subsequently, we decode it from the latent space to the original pixel space using the decoder of VQVAE.

As for the sampling in the denoising iterations, we adopt sampler proposed in \cite{huang2023decoupled} to accelerate sampling and support the two branch design. Specifically, we can generate a sample $\mathbf{x}_{t-1}$ from a sample $\mathbf{x}_t$ via:
\section{Experiment}
\label{result}
\subsection{Implementation Details}
\begin{table*}[h]
  \caption{IoU scores (\%) of HD map semantic segmentation on nuScenes dataset compared with SuperFusion.}
  \centering
  \scalebox{.9}{
  \begin{tabular}{
    l
    c
    S[table-format=2.1]
    S[table-format=2.1]
    c 
    S[table-format=2.1]
    S[table-format=2.1]
    S[table-format=2.1]
    S[table-format=2.1]
    S[table-format=2.1]
    c 
    S[table-format=2.1]
    S[table-format=2.1]
    c 
  }
    \toprule
    \multirow{2}{*}{\makecell{Method}} & \multirow{2}{*}{\makecell{Modality}} & \multicolumn{3}{c}{0-30m} & \multicolumn{3}{c}{30-60m} & \multicolumn{3}{c}{60-90m} & \multicolumn{3}{c}{\makecell{Average IoU}} \\
    \cmidrule(lr){3-14}
    & & {Divider} & {Ped.} & {Boundary} & {Divider} & {Ped.} & {Boundary} & {Divider} & {Ped.} & {Boundary}& {Divider} & {Ped} & {Boundary}\\
    \midrule
    SuperFusion & {C+L} & 47.9 & 37.4 & 58.4 & 35.6 & 22.8 & 39.4 & 29.2 & 12.2 & 28.1 & 38.0 & 26.2 & 42.7\\
    \midrule
    SuperFusion* & {C+L} & 50.9 & 39.6 & \textbf{60.6} & 38.8 & 22.8 & 42.6 & 29.7 & 13.7 & \textbf{30.3} & 40.4 & 27.4 & \textbf{44.7}\\
    
    +DiffMap & C+L & \textbf{54.3} & \textbf{41.4} & 59.5 & \textbf{43.3} & \textbf{29.1} & \textbf{43.1} & \textbf{31.3} & \textbf{21.2} & 30.1 & \textbf{43.5} & \textbf{31.9} & 44.5 \\
    \bottomrule
  \end{tabular}}
  \label{Table:IoU_result}
\end{table*}

\begin{table*}[h] 
  \centering
  \caption{mAP scores (\%) of HD map semantic segmentation on nuScenes dataset compared with SuperFusion.}
  \scalebox{.9}{
  \begin{tabular}{
    l
    c
    S[table-format=2.1]
    S[table-format=2.1]
    c 
    S[table-format=2.1]
    S[table-format=2.1]
    c 
    S[table-format=2.1]
    S[table-format=2.1]
    c 
    S[table-format=2.1]
    S[table-format=2.1]
    c 
  }
    \toprule
    \multirow{2}{*}{\makecell{Method}} & \multirow{2}{*}{\makecell{Modality}} & \multicolumn{3}{c}{0-30m} & \multicolumn{3}{c}{30-60m} & \multicolumn{3}{c}{60-90m} & \multicolumn{3}{c}{\makecell{mAP}} \\
    \cmidrule(lr){3-14}
    & & {Divider} & {Ped} & {Boundary} & {Divider} & {Ped.} & {Boundary} & {Divider} & {Ped.} & {Boundary}& {Divider} & {Ped.} & {Boundary}\\
    \midrule
    SuperFusion & {C+L} & 33.2 & 26.4 & 58.0 & 30.7 & 18.4 & 52.7 & 24.1 & 10.7 & 38.2 & 29.7 & 19.2 & 50.1\\
    \midrule
    SuperFusion* & {C+L} & 36.5 & 30.0 & \textbf{61.3} & 32.3 & 20.3 & \textbf{55.9} & 24.2 & 11.9 & \textbf{40.6} & 31.5 & 21.5 & \textbf{53.1}\\
    
         +DiffMap & C+L & \textbf{45.5} & \textbf{34.1} & 59.0 & \textbf{40.1} & \textbf{25.1} & 54.5 & \textbf{29.2} &\textbf{16.4} & 40.1 & \textbf{38.8} & \textbf{25.9} & 51.6 \\
    \bottomrule
  \end{tabular}}
  \label{Table:mAP_result}
\end{table*}

\begin{table*}[!ht]
  \centering
  \caption{Quantitative analysis of map segmentation compared with HDMapNet.}
  \scalebox{1.0}{
  \begin{tabular}{
    l
    c
    | 
    S[table-format=2.1]
    S[table-format=2.1]
    S[table-format=2.1]
    S[table-format=2.1]
    | 
    S[table-format=2.1]
    S[table-format=2.1]
    S[table-format=2.1]
    S[table-format=2.1]
  }
    \noalign{\vskip 2pt}\hline\noalign{\vskip 2pt}
    Method & Modality & {Divider} & {Ped.} & {Boundary} & {mIOU} & {Divider} & {Ped.} & {Boundary} & {mAP} \\ \noalign{\vskip 2pt}\hline\noalign{\vskip 2pt}
    HDMapNet* & C & 41.6  & 21.5  & 41.6  & 34.9  & 23.1  &  18.1 & \textbf{47.7}  &  29.6 \\
    
        +DiffMap & \multirow{2}{*}{C}   & \textbf{42.1}  & \textbf{23.9} & \textbf{42.2}  & \textbf{36.1}  &  \textbf{26.1} & \textbf{19.2}  & 45.4   & \textbf{30.2}\\ 
    
    \midrule
    HDMapNet* & C+L &  48.8 & \textbf{34.5}  &   59.6&  47.6 &  28.2 &  27.1 & \textbf{55.2} & 36.8\\ [3pt]
    +DiffMap &C+L &  \textbf{54.3} & 34.4  & \textbf{60.7}  &  \textbf{49.8} & \textbf{34.5}  &  \textbf{30.4} & 51.3 & \textbf{38.7}\\ 
    \bottomrule
    %
  \end{tabular}}
  \label{Table:modality}
\end{table*}

\begin{table*}[!ht]
  \centering
  \caption{Quantitative analysis in different conditions compared with HDMapNet (C+L).}
  \scalebox{1.0}{
  \begin{tabular}{
    l
    c
    | 
    S[table-format=2.1]
    S[table-format=2.1]
    S[table-format=2.1]
    S[table-format=2.1]
    | 
    S[table-format=2.1]
    S[table-format=2.1]
    S[table-format=2.1]
    S[table-format=2.1]
  }
    \noalign{\vskip 2pt}\hline\noalign{\vskip 2pt}
    Method & Condition & {Divider} & {Ped.} & {Boundary} & {mIOU} & {Divider} & {Ped.} & {Boundary} & {mAP} \\ \noalign{\vskip 2pt}\hline\noalign{\vskip 2pt}
    HDMapNet* & Cloudy & 54.7  & 32.7  & 61.6  & 49.7  & 29.8  &  26.9 & \textbf{55.1}  &  37.3 \\
    
    +DiffMap & Cloudy   & \textbf{58.0}  & \textbf{35.0} & \textbf{63.3}  & \textbf{52.1}  &  \textbf{35.6} & \textbf{30.3}  & 52.2   & \textbf{39.4}\\ 

    \midrule
    HDMapNet* & Rainy &  47.7 & 27.8  &   49.6& 41.7 &  27.6 &  25.1 & \textbf{52.3} & 35.0\\
    +DiffMap &Rainy &  \textbf{50.4} & \textbf{28.5} & \textbf{49.2}  & \textbf{42.7}  & \textbf{30.7}  &  \textbf{27.9} & 47.2 & \textbf{35.3}\\ 
    
    \midrule
    HDMapNet* & Night &  50.5 & \textbf{20.6}  &  60.7& \textbf{43.9}  &  27.5 &  8.2 & \textbf{62.0} & 32.6\\ [3pt]
    +DiffMap &Night &  \textbf{53.7} & 14.5  & \textbf{61.0}  &  43.1 & \textbf{30.4}  &  \textbf{11.5} & 56.8 & \textbf{32.9}\\

    \bottomrule

  \end{tabular}}
  \label{Table:condition}
. 
\end{table*}
\subsubsection{Dataset}
We validate our DiffMap on the nuScenes dataset\cite{caesar2020nuscenes}. As a popular benchmark in autonomous driving, the nuScenes dataset includes multi-view images and point clouds of 1,000 scenes, with 700 scenes allocated for training, 150 for validation, and 150 for testing. The nuScenes dataset also has annotated HD map semantic labels, which are highly suitable for our task as well. Besides training the main model, we rasterize the ego map data into segmentation maps during the pre-training of the autoencoder.

\subsubsection{Architecture}
We use ResNet-101\cite{he2016deep} as the backbone for our camera branch and PointPillars\cite{lang2019pointpillars} as our LiDAR branch backbone. The Segmentation head in baseline is a FCN network based on ResNet-18. For the autoencoder, we adopt VQVAE, which is pre-trained on the nuScenes segmentation maps dataset, to extract map features and compress map into the basic latent space. 
UNet is used to construct the diffusion network.

\subsubsection{Training Details}
We conducted our model training and baseline reproduction using 8 NVIDIA RTX A6000 GPUs. 
Firstly, we train the VQVAE model for 30 epochs using the AdamW optimizer. The learning rate scheduler employed is LambdaLR, which gradually reduces the learning rate with an exponential decay pattern using a decay factor of 0.95. The initial learning rate is set to $5e^{-6}$, and the batch size is 8. Then, we train diffusion model from scratch for 30 epochs using AdamW optimizer and an initial learning rate of $2e^{-4}$. We adopt MultiStepL scheduler which adjusts the learning rate in stages based on specified milestone time points $(0.7, 0.9, 1.0)$ and a scaling factor $1/3$ during different stages of training. 
We set the BEV segmentation result and voxelize the LiDAR point cloud at a resolution of $0.15 \text{m}$. HDMapNet's detection range is $[-30\text{m}, 30\text{m}] \times [-15\text{m}, 15\text{m}] $, so the corresponding BEV map size is $400 \times 200$, while Superfusion uses $[0\text{m}, 90\text{m}] \times [-15\text{m}, 15\text{m}]$ and gets the results of $600 \times 200$. 
Due to the dimension constraints of the LDM (downsampled by a factor of 8x in VAE and 8x in UNet), we pad the size of semantic ground truth map into a multiple of 64.

\subsubsection{Inference Details}
The prediction results are obtained by performing the denoising process $20$ steps on the noisy maps under the condition of current BEV features. Considering the stochastic nature of the generative model, we perform the sampling process $3$ times, and use the mean value of sampling results as the final prediction result. We test the HDMapNet(C) with DiffMap module on one A6000 GPU, the inference FPS is 0.32.
\subsection{Evaluation Metrics}
We evaluate two target sub-tasks, namely map semantic segmentation and instance detection, and introduce metrics specific to each task. 
Following the methodology outlined in HDMapNet and Superfusion, our evaluation primarily focuses on three static map elements: lane boundary, lane divider, and pedestrian crossing.

\subsubsection{Semantic Metrics}

We use IoU as the Eulerian metrics between the predicted HD Map $M_1$ and the ground-truth HD Map $M_2$, which is given by :
\begin{equation}
\text{IoU}(M_1, M_2) = \frac{M_1 \cap M_2}{M_1 \cup M_2}    
\end{equation}

\subsubsection{Instance Metrics}
To evaluate spacial distances between predicted curves and ground-truth curves, We use Chamfer distance(CD), which is defined as:
\begin{equation}
    \text{CD}_{\text{dir}}(C_1, C_2) = \frac{1}{C_1} \sum_{x \in C_1} \mathop{\min}_{y \in C_2} \|x - y\|_2 
\end{equation}

where $C_1$ and $C_2$ are a set of points on the predicted curve and ground-truth curve. The typical Chamfer distance(CD) is bi-directional. And the CD between the predicted curve and the ground-truth curve is given by: 
\begin{equation}
    \begin{split}
    \text{CD} &= \text{CD}_\text{pred} + \text{CD}_\text{gt} \\
       &= \text{CD}_\text{dir}(C_1, C_2) + \text{CD}_\text{dir}(C_2, C_1)
    \end{split}
\end{equation}

Then we use AP (Average Precision) to measure the instance detection distance, which is defined as:
\begin{equation}
    \text{AP} = \frac{1}{10} \sum_{r \in {0.1, 0.2, ..., 1.0}} \text{AP}_r
\end{equation}

where $\text{AP}_r$ is the precision at recall $r$. Following  \cite{dong2022superfusion}, we simultaneously use both CD and IoU to select true positive instances. The instance is considered as a true positive if and only if the IoU is above a certain threshold and the CD is below another threshold. Specifically, we set the threshold of IoU as 0.1 and threshold of CD as 1.0m. 

\subsubsection{Evaluation on multiple intervals}
To better explore the effectiveness of DiffMap in long-range detection, we further split the ground truth into three intervals: 0-30m, 30-60m, and 60-90m to compare with SuperFusion fairly. 
Subsequently, within these three intervals, we calculate the IoU, CD, and AP for different methods, aiming for a comprehensive evaluation of the detection results.

\begin{figure*}[t]
    \centering
    \includegraphics[width=0.88\textwidth]{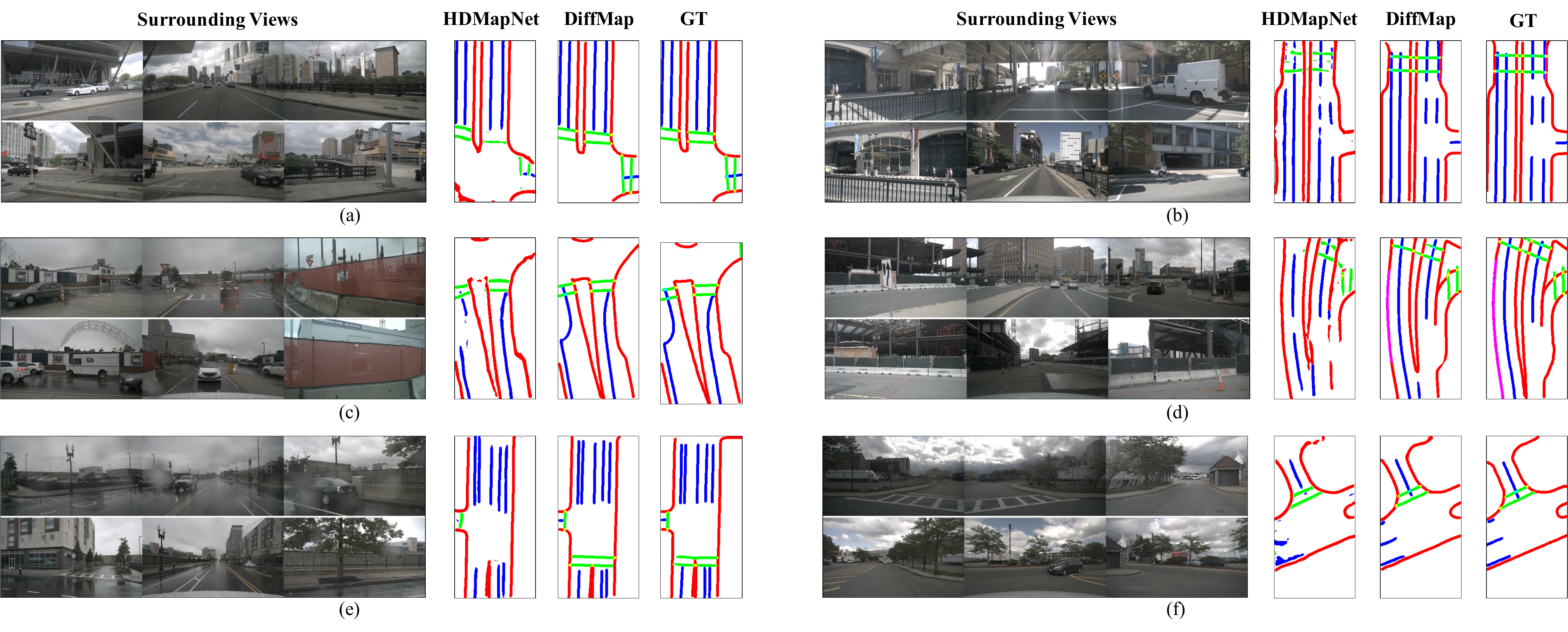}
    \caption{ \centering{\textbf{Qualitative results on short range map segmentation.}} }
    \label{fig:vis}
\end{figure*}

\begin{figure*}[t]
    \centering
    \includegraphics[width=0.88\textwidth]{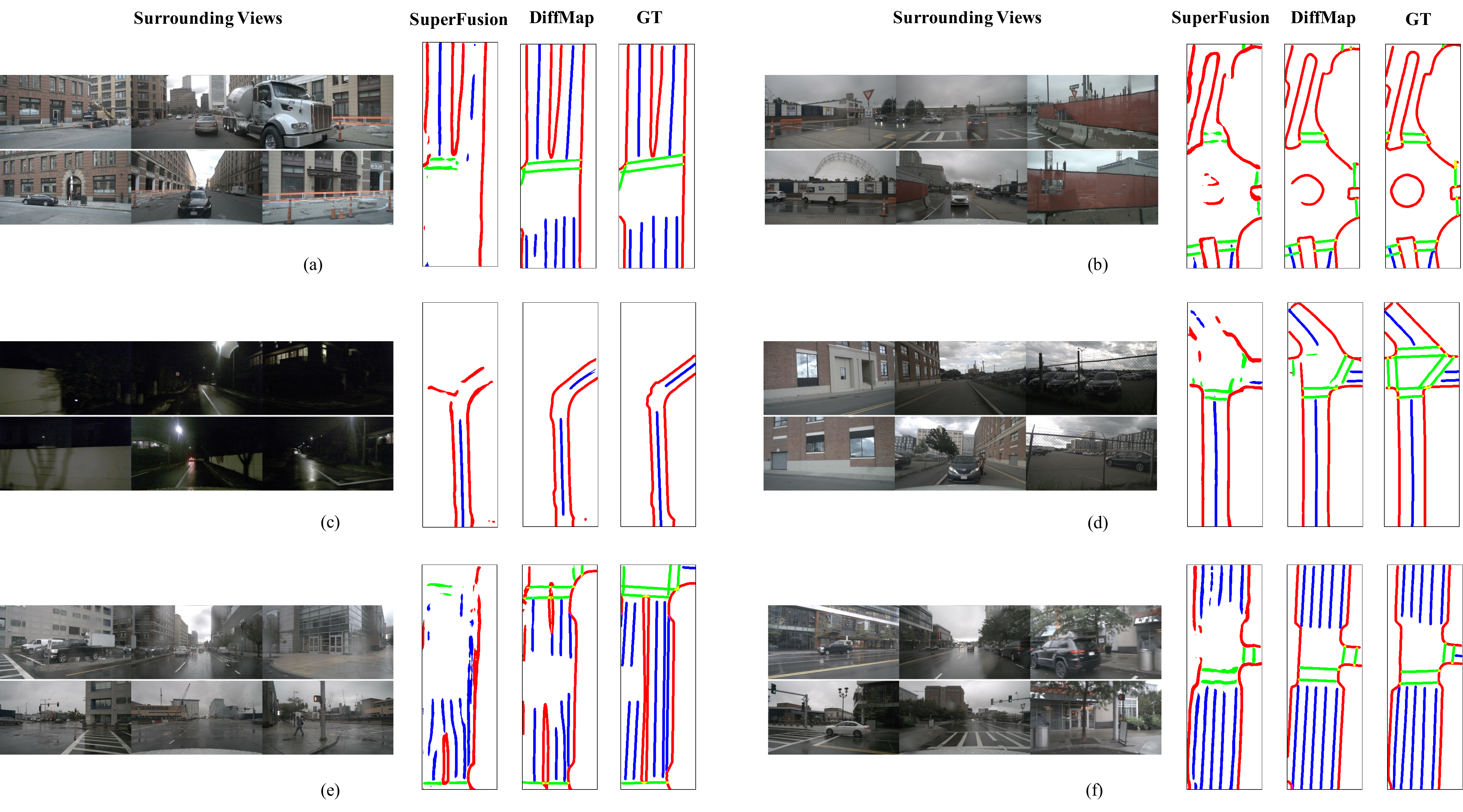}
    \caption{\textbf{Qualitative results on long range map segmentation:} Diffmap is capable of capturing the structured prior and achieves the best segmentation results. The results demonstrate the effective restoration of parallel shapes for pedestrian crossings, smoothness and continuity of dividers, and shape complementation for boundaries.}
    \label{fig:vis_long}
\end{figure*}
\subsection{Evaluation Results}
 To facilitate better comparison, we reproduce the predominant methods for map segmentation like HDMapNet and SuperFusion algorithms on nuScenes dataset. 
 
 Compared with Superfuion, Table \ref{Table:IoU_result} shows the comparisons of the IoU scores of semantic map segmentation (where * means the model reproduced by us). Our DiffMap shows a significant improvement on all intervals and has best results, especially in lane divider and pedestrian crossing. 
 As shown in Table \ref{Table:mAP_result}, our method also demonstrates a significant improvement in terms of average precision (AP), verifying the effectiveness of our DiffMap. 
We also incorporate our DiffMap paradigm into HDMapNet and get the result shown in Table \ref{Table:modality}. 
We can observe that whether it is camera-only or camera-LiDAR fusion method, DiffMap improves the performance of HDMapNet. 
They illustrate our effectiveness on all types of segmentation tasks, including both long and short range detection.

Besides we conduct experiments on cloudy, rainy and night conditions based on HDMapNet with modality C+L, the results are shown in Table \ref{Table:condition}.

However, we also find that for boundaries, DiffMap does not perform well, and we conjecture that this is because the shape structure of boundaries has many unpredictable torsion, thereby causing difficulties in capturing the a priori structural features and although there is an obvious completion correction effect on the shape, the true position is still biased.

\subsection{Ablation Analysis}
Table \ref{Ablation: downsampling} shows the impact of different downsampling factors in VQVAE on the detection results. We analyze the behavior of our DiffMap with different downsampling factors of 4, 8, 16. And We can observe that results are best when downsampling factor is set to 8x.

\begin{table}[!h]
  \centering
  \caption{Ablation study on the different VAE downsampling factors.}
  \renewcommand\arraystretch{1.5}
  \scalebox{0.8}{
  \begin{tabular}{
    c
    | 
    c
    S[table-format=2.1]
    S[table-format=2.1]
    S[table-format=2.1]
    | 
    S[table-format=2.1]
    S[table-format=2.1]
    S[table-format=2.1]
    S[table-format=2.1]
  }
    \noalign{\vskip 2pt}\hline\noalign{\vskip 2pt}
    Method &  {Divider} & {Ped.} & {Boundary} & {mIOU} & {Divider} & {Ped.} & {Boundary} & {mAP} \\ \noalign{\vskip 2pt}\hline\noalign{\vskip 2pt}
    4× &  51.4  &  \textbf{35.1} &  59.2 & 48.6  & 30.1  & 27.8  & 49.3  & 35.7  \\ \noalign{\vskip 2pt}\hline\noalign{\vskip 2pt}
    8× &   \textbf{54.3} & 34.4  & \textbf{60.7}  &  \textbf{49.8} & \textbf{34.5} &  \textbf{30.4} & \textbf{51.3} & \textbf{38.7}  \\ \noalign{\vskip 2pt}\hline\noalign{\vskip 2pt}
    16× & 51.8 & 30.4  &  57.5 & 46.6  &  24.9 &  20.9 &  48.1 & 31.3  \\ \noalign{\vskip 2pt}\hline\noalign{\vskip 2pt}
  \end{tabular}}
  \label{Ablation: downsampling}
  
 \end{table}

 Besides, we measure the effect of removing the instance-related prediction module on the model, as shown in Table \ref{Ablation: instance}. Experiments show that adding this prediction further improves the IOU.

 \begin{table}[ht]
  \centering
  \caption{Ablation study on whether add the instance-related prediction module.} 
  \label{Ablation: instance}
  \begin{tabular}{
    >{\centering\arraybackslash}m{2cm} 
    | 
    c
    S[table-format=2.1]
    S[table-format=2.1]
    S[table-format=2.1]
  }
    \noalign{\vskip 2pt}\hline\noalign{\vskip 2pt}
    Method &  {Divider} & {Ped.} & {Boundary} & {mIOU}  \\ \noalign{\vskip 2pt}\hline\noalign{\vskip 2pt}
    w/o ins &  52.9  & \textbf{34.6}  &  60.1 &   49.2\\ \noalign{\vskip 2pt}\hline\noalign{\vskip 2pt}
    w ins & \textbf{54.3}   & 34.4  & \textbf{60.7}  &  \textbf{49.8} \\ \noalign{\vskip 2pt}\hline\noalign{\vskip 2pt}
  \end{tabular}
\end{table}

\subsection{Visualization}
Figure \ref{fig:vis} illustrates the comparison between DiffMap and the baseline (HDMapNet-fusion) in complex scenarios and Figure \ref{fig:vis_long} illustrates the performance of DiffMap in long-range map segmentation. It is evident that the segmentation results of the baseline disregard the shape properties and coherence within elements. 
In contrast, DiffMap demonstrates the ability to address these problems, resulting in segmentation outputs that align well with the specifications of the map. 


\section{CONCLUSIONS}
\label{conclusion}

In this paper, we introduce DiffMap, a novel approach that utilizes latent diffusion model to learn the structured priors of maps, thereby enhancing the traditional map segmentation models. Our method can serve as an auxiliary tool for any map segmentation model and its predicted results have significant improvements in both short and long range detection scenarios.

\textbf{Limitation and Future work:} Looking ahead, we are eager to explore the potential for direct application of diffusion models in the construction of vectorized maps. Additionally, the current model sampling latency is excessive, failing to meet real-time requirements. Therefore, further acceleration is necessary to achieve timely performance.

\bibliographystyle{IEEEtran}
\bibliography{ref}

\end{document}